\documentclass[conference]{IEEEtran}
\IEEEoverridecommandlockouts
\usepackage{cite}
\usepackage{amsmath,amssymb,amsfonts}
\usepackage{algorithmic}
\usepackage{graphicx}
\usepackage{textcomp}
\usepackage{xcolor}
\usepackage{caption}
\usepackage{subcaption}
\def\BibTeX{{\rm B\kern-.05em{\sc i\kern-.025em b}\kern-.08em
    T\kern-.1667em\lower.7ex\hbox{E}\kern-.125emX}}
\begin{document}

\title{Shutter, the Robot Photographer: Leveraging Behavior Trees for Public, In-the-Wild Human-Robot Interactions\\
\thanks{ This work was partially supported by the National Science
Foundation (NSF), Grant No. (IIS-1924802). 
The authors are thankful to many students who have contributed to Shutter, including B. Lyng-Olsen, T. Adamson, V. Del Carpio, Y. Milkessa, E. Gorevoy, A. Narcomey, Q. Zhang, and S. Gillet.}
}

\author{
\IEEEauthorblockN{Alexander Lew}
\IEEEauthorblockA{\textit{Computer Science} \\
\textit{Yale University}\\
New Haven, CT \\
a.lew@yale.edu}
\and
\IEEEauthorblockN{Sydney Thompson}
\IEEEauthorblockA{\textit{Computer Science} \\
\textit{Yale University}\\
New Haven, CT \\
sydney.thompson@yale.edu}
\and
\IEEEauthorblockN{Nathan Tsoi}
\IEEEauthorblockA{\textit{Computer Science} \\
\textit{Yale University}\\
New Haven, CT \\
nathan.tsoi@yale.edu}
\and
\IEEEauthorblockN{Marynel V{\'a}zquez}
\IEEEauthorblockA{\textit{Computer Science} \\
\textit{Yale University}\\
New Haven, CT \\
marynel.vazquez@yale.edu}
}

\maketitle

\begin{abstract}
    Deploying interactive systems in-the-wild requires adaptability to situations not encountered in lab environments.
    Our work details our experience about the impact of architecture choice on behavior reusability and reactivity while deploying a public interactive system.
    In particular, we introduce Shutter, a robot photographer and a platform for public interaction. In designing Shutter's architecture, we focused on adaptability for in-the-wild deployment, while developing a reusable platform to facilitate future research in public human-robot interaction. 
    We find that behavior trees allow reactivity, especially in group settings, and encourage designing reusable behaviors. 
\end{abstract}

\section{Introduction}

Public interactive agents face unique challenges that do not occur in controlled environments. In-the-wild deployment requires robustness, both to the environment and to users, because public interactions frequently have unexpected circumstances that demand agile adaptation from the agent, and may not have experts nearby to correct errors. A common method of handling these varied situations is by defining specific robot responses for these cases \cite{directions-robot-2014}.

The complexity of public spaces has motivated the development of better methods for perception and interaction. For instance, group detection \cite{swofford2020improving} is an important capability for improving robot activity in social environments. As the number of modalities and capability of these agents increase, orchestrating robot activity in a simple yet robust manner is a critical step in designing compelling interactions. The underlying decision-making architecture of a robot shapes the design of the robot's interactive actions, with specific regard to the modularity and reusability of those actions. 

In this work, we describe Shutter, a robot photographer, and our experience in implementing a behavior architecture to enable responsive and reusable interaction with its environment. The primary contributions of this work are the introduction of Shutter as a public interaction platform and a discussion of the lessons learned while designing a modular and reactive system for in-the-wild deployments.

\begin{figure}[t]
\centering
 \includegraphics[width=0.5\textwidth]{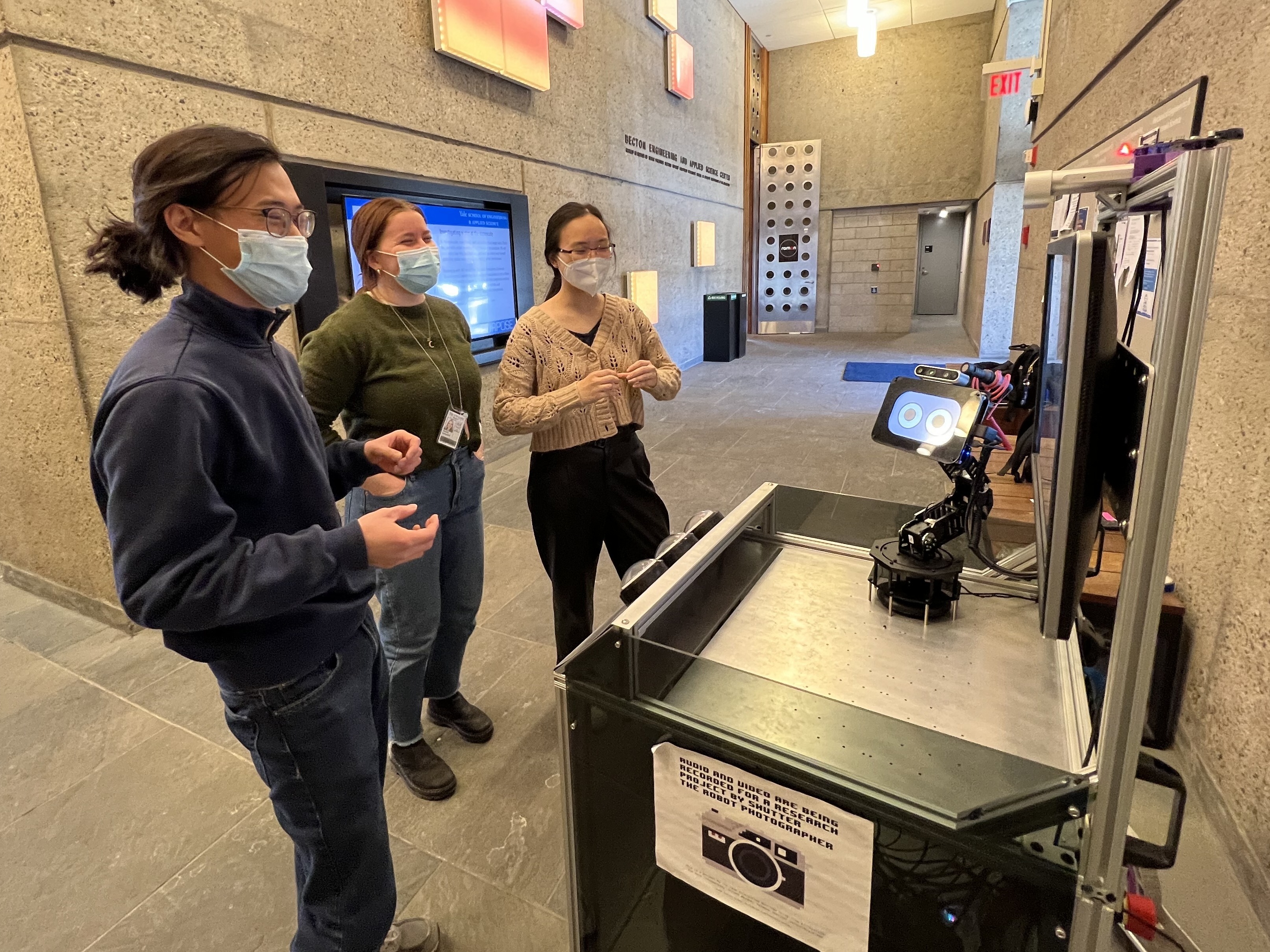}
 \caption{Shutter, a robot photographer platform designed for public interactions.}
 \label{fig:shutter}
\end{figure}
 


\section{Related Work}
\begin{figure*}
     \centering
     \begin{subfigure}{0.49\textwidth}
         \centering
         \includegraphics[width=.89\textwidth]{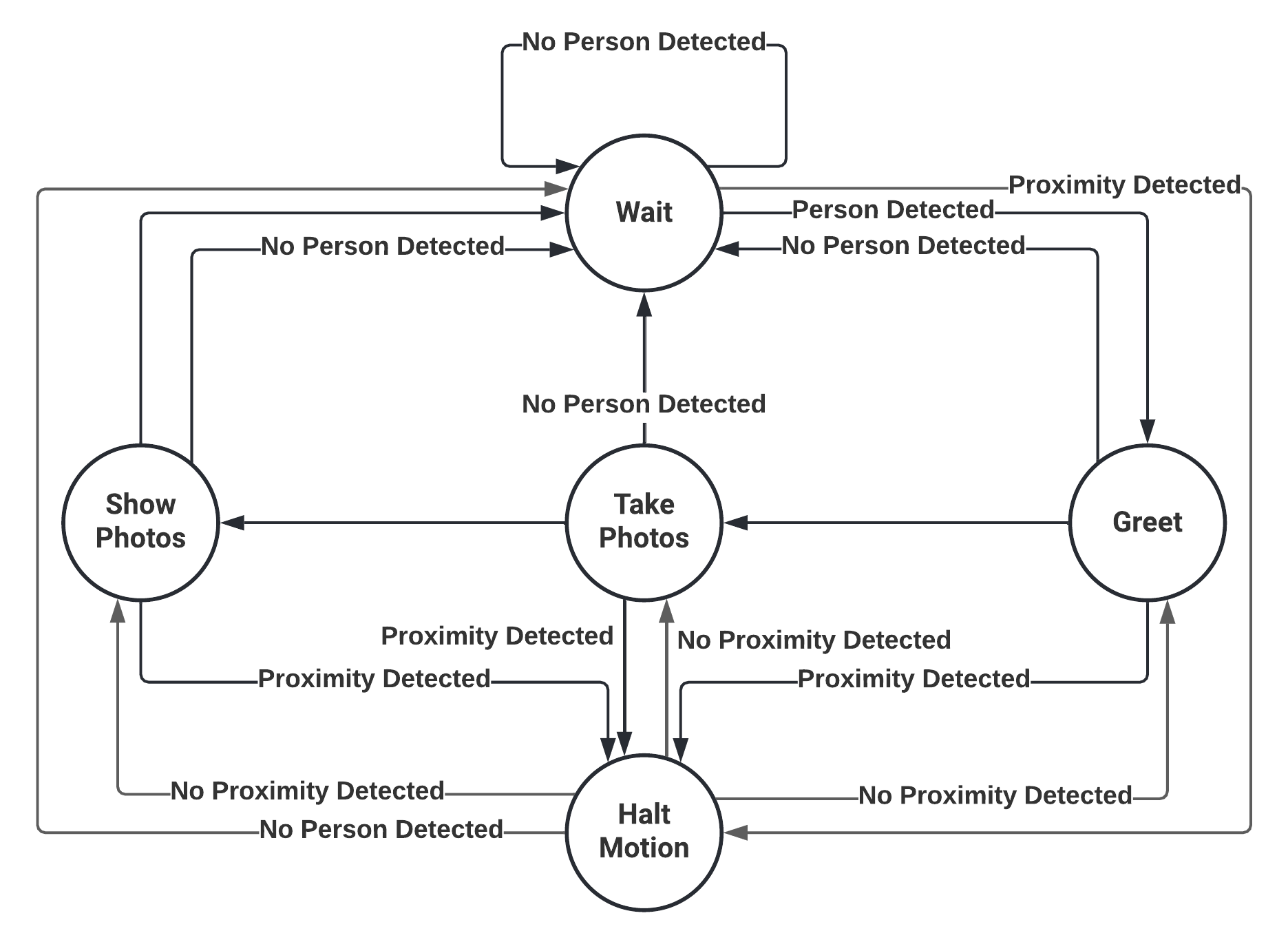}
         \caption{Finite state machine representing Shutter's basic interaction.}
         \label{fig:basic_fsm}
     \end{subfigure}
     \hfill
     \begin{subfigure}{0.49\textwidth}
         \centering
         \includegraphics[width=.8\textwidth]{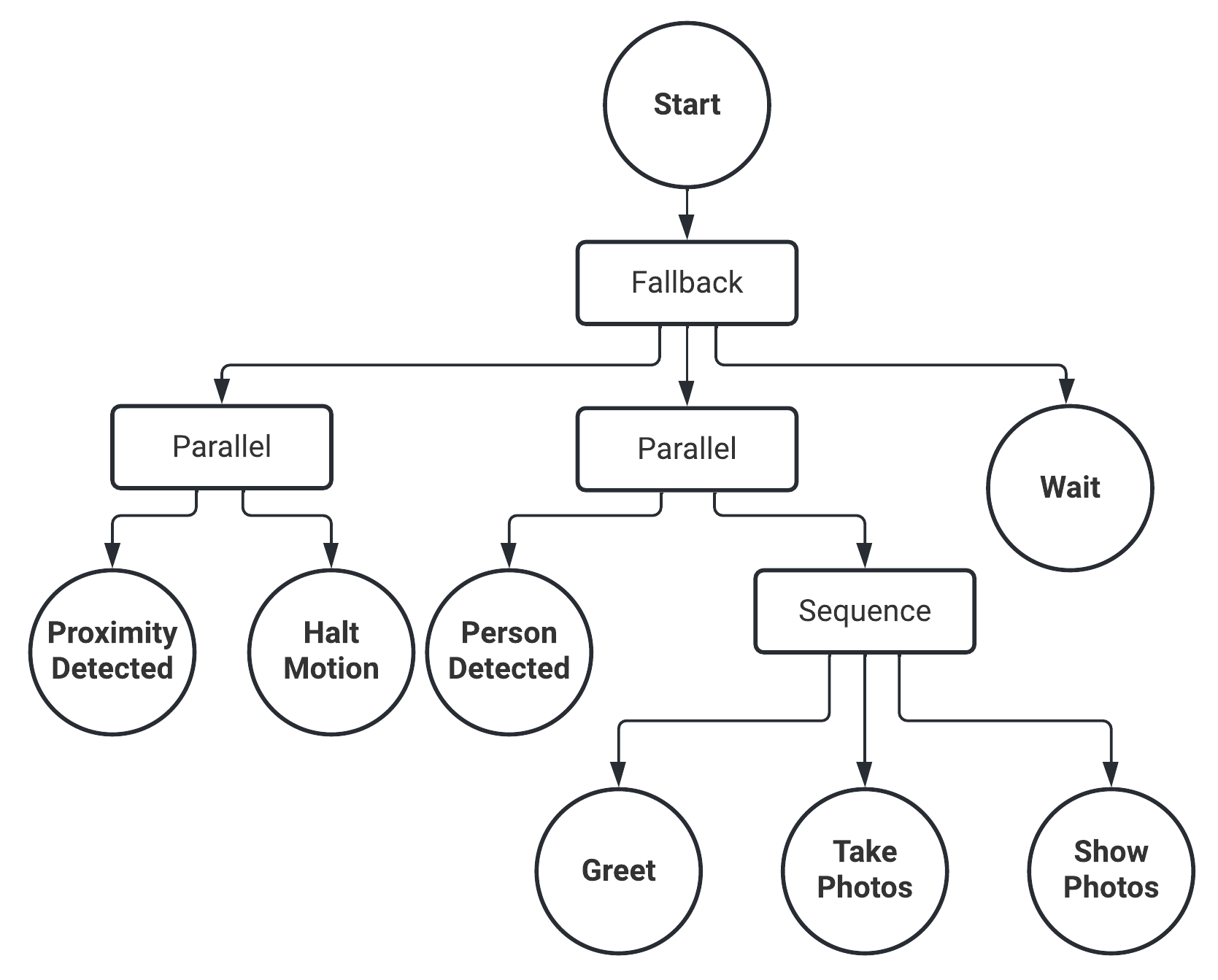}
         \caption{Behavior tree representing Shutter's basic interaction.}
         \label{fig:basic_bt}
     \end{subfigure}
     \caption{Shutter's behavior architecture implemented with both a finite state machine and a behavior tree.}
    \label{fig:basic}
\end{figure*}

Several control architectures have been proposed for social robots deployed in public environments, including cognitive architectures, finite state machines, and behavior trees. Cognitive architectures both model and select interaction responses \cite{10.5898/JHRI.2.1.Trafton,doi:10.1177/0278364917690592}. Finite state machines are a classic method of specifying robot activity and have been used in a variety of domains, including guiding museum tours \cite{del2019lindsey} and search-and-rescue \cite{de2014parallel}. The finite state machine architecture is comprised of states that define actions and transitions that represent reactions to observations or user input. Transition conditions define which transition is taken from each state. See Fig. \ref{fig:basic_fsm} for an example. Transitions between states could be initiated by the user's input via buttons, a touchscreen interface\cite{bauer-2009}, tactile responses \cite{kanda-2010}, verbal responses from the user \cite{directions-robot-2014}, or the robot's successful completion of an action \cite{wang-2018}.

There are many examples of state machines utilized for public interactive robots. 
The Directions Robot  \cite{directions-robot-2014} employed a finite-state based dialog manager to control conversation actions conditioned on scene analysis in order to give users navigation directions. In a shopping mall deployment of Robovie \cite{kanda-2010}, robot activity was coordinated by a control module according to contextual information and pre-implemented episode rules. These works also explore the challenges of deployments in public environments, such as managing unexpected user actions and observing spontaneous engagement and disengagement. In social robotics, Bauer et al. \cite {bauer-2009} implemented a finite state machine that defined an interaction to ask users for directions to a goal position.  A finite state machine is also utilized by \cite{wang-2018} to orchestrate the high-level activity of a mobile robot that serves as a tour guide.

Behavior trees have also been employed previously in designing public interactions. A behavior tree is another structure for specifying the execution of actions that an autonomous agent can perform \cite{bt-book-2018}, \cite{iovino2020survey}.
Leaf nodes represent the behaviors that the agent can execute. Generating speech, gathering sensor data, and updating an internal data structure are all examples of behaviors. The order in which behaviors are executed is determined by the internal, or control flow, nodes. Fallback, parallel, and sequence are the main types of control flow nodes. Traversal of the behavior tree proceeds at regular intervals, or tick events, from the root. At each received tick, a behavior returns a status of success, running, or failure, which the parent control flow node uses to determine the next node to receive a tick event. Fig. \ref{fig:basic_bt} has examples of the three main types of internal nodes in a behavior tree that defines a basic photo-taking interaction with Shutter.

In \cite{coronado-2019}, behavior trees are evaluated as a framework to script child-robot interactions in-the-wild. A major difference between this work and our proposal is the robot's degree of group awareness, as the reactive behaviors in \cite{coronado-2019} assume continual user presence.
\section{Shutter, a Robot Photographer}

Our social robot, Shutter, is a robot photographer and was developed as a platform for public interactions; we intended it to be used in a hallway, foyer, or gallery setting for many kinds of photo-taking interactions. To this end, we value reusability in Shutter's architecture. While the primary purpose of the robot is to take photos, we designed Shutter to support many kinds of interactions. For our first implementation, Shutter detects potential interactants, takes posed photos, and shows the photos to the user. 

\subsection{Hardware}
Fig. \ref{fig:shutter} depicts our robotic platform. Shutter is positioned on a cart that allows deployment to a variety of indoor locations. The robot is a Trossen Robotics WidowX Robot Arm, with an added facial display. An Intel RealSense RGB-D camera is also mounted on the robot's head, providing egocentric perception. An Azure Kinect sensor on the cart provides a third-person perspective of the environment and performs body tracking. The Kinect is mounted above a large display screen that serves as a complementary visual interface for users. A loudspeaker positioned underneath the robot plays audio, including generated speech. Users can provide feedback to Shutter, such as choosing options from a menu, by pressing any of the three illuminated buttons on the front of the cart.


\subsection{Software}
We created custom software for controlling low-level actions, like motion and rendering facial expressions, and for implementing the higher-level control architecture required for specifying interactions. Shutter also leverages existing libraries, e.g., to extract the Azure Kinect's body tracking data. To facilitate the creation of more public interactive systems, the source code for Shutter's control architecture and action definitions is publicly available.






Group awareness is key for creating adaptive responses for Shutter. A group detection algorithm \cite{swofford2020improving} clusters people into social groups, including the social group containing Shutter. In practice, detecting groups allows for dynamic changes to the interaction. For example, Shutter currently detects the number of interactants and adapts the greeting for the group size. If there were 3 people interacting with Shutter, as in Fig. \ref{fig:shutter}, Shutter would greet the group by saying "Would you like me to take a photo of the three of you?".
As in modeling the groups themselves \cite{swofford2020improving}, the number of people in a social group is not known a priori, so more involved group-aware actions from Shutter will need to adapt to this change in group size. Additionally, group size can change during the interaction. Thus, when evaluating software architectures for Shutter, we prioritized agility around group size and current interactants.


\section{Architecture Comparison}

When defining Shutter's software architecture to realize the intended interaction, we considered two approaches: finite state machines and behavior trees. While both have similar capabilities in practice, the structure imposed by each architecture changes the way actions can be reused and how easily reactive behaviors are implemented.

\subsection{Basic Shutter Interaction}
An interaction begins when Shutter's Kinect sensor detects at least one person nearby. The robot introduces itself and asks ``Would you like me to take your photo?" or ``Would you like me to take a photo of the $n$ of you?" if there are $n > 1$ users. After receiving the user or users' consent, Shutter tells the interactant(s) that it is about to take their photo and takes three photos with the onboard RealSense camera. The photos are shown on the display as Shutter praises the photos with remarks like ``You look great in this photo." 

\subsection{Finite State Machine Architecture}

Initially, we implemented Shutter's interaction using a finite state machine. In our finite state machine, nodes represent high-level activity that Shutter takes, and transitions represent human responses that require a change in the robot's activity. Fig. \ref{fig:basic_fsm} shows the finite state machine. Following transitions from state to state provides a semantic understanding of Shutter's activity. A primary benefit of our finite state machine is that these explicit transitions provide transparency in Shutter's responses.

However, as the number of states of our finite state machine grew, we found that deciphering the transitions and creating reactive responses to a changing environment became much more difficult. This is because we defined behavior interactions with explicit transitions. A state like ``Halt Motion", which needed to be reachable from every point in the interaction, must therefore have multiple transitions into that state. This increased the number of transitions required in our finite state machine. Overall, we found that this overhead limited the reusability of sub-graphs from our initial finite state machine. When trying to extend our finite state machine for different interaction designs, we struggled to efficiently modify the default interactions.

\subsection{Behavior Tree Architecture}

To address our concerns about the finite sate machine, we implemented a behavior tree for Shutter with behaviors that either progress the ongoing interaction or update the interaction context. Coordinating these two categories of behaviors allowed us to achieve reactive robot control. For example, detecting a nearby person to start the basic interaction constitutes a behavior that updates the interaction context. An example of such an update is the node  ``Person Detected" in Fig. \ref{fig:basic_bt}.

A key consideration when developing the behavior tree was behavior reusability. For instance, the same proximity detection behavior that starts the interaction is reused to stop the interaction if the user is no longer detected. Subtrees defining more complex behaviors, such as the registering of button presses, are also reused. This enabled us to more easily define new interactions by connecting existing subtrees in new ways. The next section describes specific interaction scenarios where we found value in behavior trees in comparison to finite state machines.

\subsection{Motivating Examples for Using Behavior Trees}

\subsubsection{Interaction abandonment}

Unlike in a lab setting, public interactive systems must be robust to interaction abandonment by users because they may become distracted or need to leave during an interaction. For Shutter, this takes the form of having no person detected by the Kinect on its cart while an interaction is ongoing and returning to the waiting state.

Handling interaction abandonment with a finite state machine requires one of two solutions: 1) transitioning back to a starting state after a user leaves, or 2) finishing the interaction without a user present. As shown in Fig. \ref{fig:basic_fsm}, adding transitions to account for abandonment requires transitions from each state back to the waiting state. Alternatively, each state could have a time-out option that returns the finite state machine to the waiting state. However, this solution requires each state until the end of the interaction to reach the timeout. While these solutions are easily interpretable, they require a high number of additional transitions or a long reset period.

In contrast to the finite state machine, behavior trees are designed to quickly recover from such a failure case. As shown in Fig. \ref{fig:basic_bt}, if the ``Person Detected" node fails, the parent Parallel block fails, and the preceding Fallback node defaults to the waiting state. Using a behavior tree to handle interaction abandonment requires only one architecture change, adding the Parallel block, while changing the finite state machine requires an additional transition or timeout for every node.

\subsubsection{Higher-priority interruptions}
Another in-the-wild scenario that decision making architectures must reliably manage is a high-priority interruption or fallback. During a natural interaction, a condition could require the robot to take immediate, specific action. One example of this action is a safety fallback, in which the robot must interrupt its planned activity to reduce risks. Another example is pausing an interaction in the event of network connection loss. Once the prioritized condition is satisfied, however, the interaction should be resumed in order to preserve engagement. 

We implemented a motion halt fallback as a higher-priority interruption of ongoing interactions in our finite state machine and behavior tree. During Shutter's operation, a user reaching toward the moving robot arm constitutes a potential hazard. To reduce the risk of injuring a user, the robot halts its planned motion until the user moves away. 

Similar to interaction abandonment, to implement a priority fallback in a finite state machine, a transition to the higher-priority state must be defined for every other state. As the number of states needed to represent the interaction grows, the number of required transitions also increases. Furthermore, transitions from the higher-priority state to the originating state must also be defined so that the interaction can resume. Preserving this originating context represents additional overhead during the traversal and inspection of a finite state machine. 

In contrast, our behavior tree interrupts its execution by continually checking the priority condition in a similar manner to checking for a user's presence. If the priority condition is active, the motion halt continues to block tree traversal. Then, when the condition has been cleared, the interaction proceeds from the same place, i.e., where the interruption occurred. Because the traversal of the behavior tree can be controlled directly by control flow nodes and behavior statuses for each tick event, the resumption of the interaction is enacted without separately recording the context at the moment of the interruption, as required by our finite state machine implementation.

\subsection{Challenges with Behavior Trees}
Some components of an interaction are more clearly articulated with a finite state machine than with a behavior tree. Because behavior tree nodes are designed to be self-contained, even simple action patterns must be explicitly enumerated. Hence, designing efficient, correct behavior trees requires a higher degree of familiarity with the architecture than designing simple finite state machines. Button presses are an example of a knowledge barrier imposed by behavior trees. In our finite state machine, button presses intuitively represent transitions that change robot behavior. In our behavior tree, each button press is implemented as the child of a  parallel node. More sophisticated forms of user input, such as many button press combinations or timeout conditions, would demand a more complex structure.
\section{Future Work}


The design principles set forth by the behavior tree architecture motivate further work in real-world deployments. In particular, Shutter's behaviors and the trees built with those behaviors are modular and self-contained, which facilitates re-use. Owing to this re-usability, a variety of interaction scripts can be quickly developed and deployed. Augmenting Shutter's perception capabilities, such as adding a behavior for hotword detection, could enable more modes of interaction. Improving our group detection method and incorporating it into additional interactions could further improve fluency. 
For instance, Shutter could make inquiries about which nearby people should be included in the photo, or ask each interactant review the photos individually.

The deployment of multiple copies of Shutter could also be improved by defining and handling robust safety conditions and fallbacks. Reacting safely to known hazards would permit greater autonomy by reducing the need for an expert operator or supervisor.

As a relatively low-cost platform with open-source components, Shutter could be reproduced to scale data collection and user studies in-the-wild. Because the architecture of behavior trees encourages modularity, a greater variety of distinct interactions could be deployed in public spaces. Shutter taking the role of a group mediator, a purveyor of community information, or a teacher of robotic systems to the general public would all represent examples of interactions that could be specified with behavior trees that reuse many of our existing software components.


\bibliographystyle{IEEEtran}
\bibliography{IEEEabrv,references}

\end{document}